\documentclass[lettersize,journal]{IEEEtran}
\usepackage{amsmath,amsfonts}
\usepackage{algorithmic}
\usepackage{algorithm}
\usepackage{array}
\usepackage{textcomp}
\usepackage{stfloats}
\usepackage{url}
\usepackage{verbatim}
\usepackage{graphicx}
\usepackage{cite}
\usepackage{booktabs}
\usepackage{color, colortbl} 
\usepackage{multirow, multicol, makecell} 
\usepackage{pifont}
\usepackage{threeparttable}
\usepackage{subfigure}
\usepackage{soul}
\usepackage[dvipsnames]{xcolor}

\definecolor{ztfgray}{rgb}{0.85, 0.85, 0.85} 
\newcommand{\CC}[1]{\cellcolor{ztfgray!#1}}

\begin{document}

\title{CAS-ViT: Convolutional Additive Self-attention Vision Transformers for Efficient Mobile Applications}

\author{Tianfang~Zhang, Lei~Li, Yang~Zhou, Wentao~Liu, Chen~Qian, Jenq-Neng~Hwang, Xiangyang~Ji,~\IEEEmembership{Member, IEEE}
\thanks{\textit{Corresponding author: Xiangyang~Ji}.}
\thanks{Tianfang~Zhang is with the SenseTime Research and the Department of Automation, Tsinghua University, Beijing, 100190, China (e-mail: sparkcarleton@gmail). Lei~Li is with the Department of Electrical and Computer Engineering, University of Washington and Department of Computer Science, University of Copenhagen (e-mail: lilei@di.ku.dk). Yang~Zhou, Wentao~Liu and Chen~Qian are with the SenseTime Research (e-mail: zhouyang@sensetime.com, liuwentao@sensetime.com, qianchen@sensetime.com). Jenq-Neng~Hwang is with the Department of Electrical and Computer Engineering, University of Washington (e-mail: hwang@uw.edu). Xiangyang~Ji is with the Department of Automation, Tsinghua University, Beijing, 100190, China (e-mail: xyji@tsinghua.edu.cn)}
}

\markboth{Journal of \LaTeX\ Class Files,~Vol.~14, No.~8, August~2021}%
{Shell \MakeLowercase{\textit{et al.}}: A Sample Article Using IEEEtran.cls for IEEE Journals}


\maketitle

\begin{abstract}
Vision Transformers (ViTs) mark a revolutionary advance in neural networks with their token mixer's powerful global context capability. 
However, the pairwise token affinity and complex matrix operations limit its deployment on resource-constrained scenarios and real-time applications, such as mobile devices, although considerable efforts have been made in previous works.
In this paper, we introduce CAS-ViT: Convolutional Additive Self-attention Vision Transformers, to achieve a balance between efficiency and performance in mobile applications.
Firstly, we argue that the capability of token mixers to obtain global contextual information hinges on multiple information interactions, such as spatial and channel domains.
Subsequently, we propose Convolutional Additive Token Mixer (CATM) employing underlying spatial and channel attention as novel interaction forms. This module eliminates troublesome complex operations such as matrix multiplication and Softmax.
We introduce Convolutional Additive Self-attention(CAS) block hybrid architecture and utilize CATM for each block. And further, we build a family of lightweight networks, which can be easily extended to various downstream tasks.
Finally, we evaluate CAS-ViT across a variety of vision tasks, including image classification, object detection, instance segmentation, and semantic segmentation.
Our M and T model achieves 83.0\%/84.1\% top-1 with only 12M/21M parameters on ImageNet-1K. 
Meanwhile, throughput evaluations on GPUs, ONNX, and iPhones also demonstrate superior results compared to other state-of-the-art backbones. 
Extensive experiments demonstrate that our approach achieves a better balance of performance, efficient inference and easy-to-deploy.
Our code and model are available at: \url{https://github.com/Tianfang-Zhang/CAS-ViT}

\end{abstract}

\section{Introduction}

\begin{figure*}[!t]
  \centering
   \includegraphics[width=0.95\linewidth]{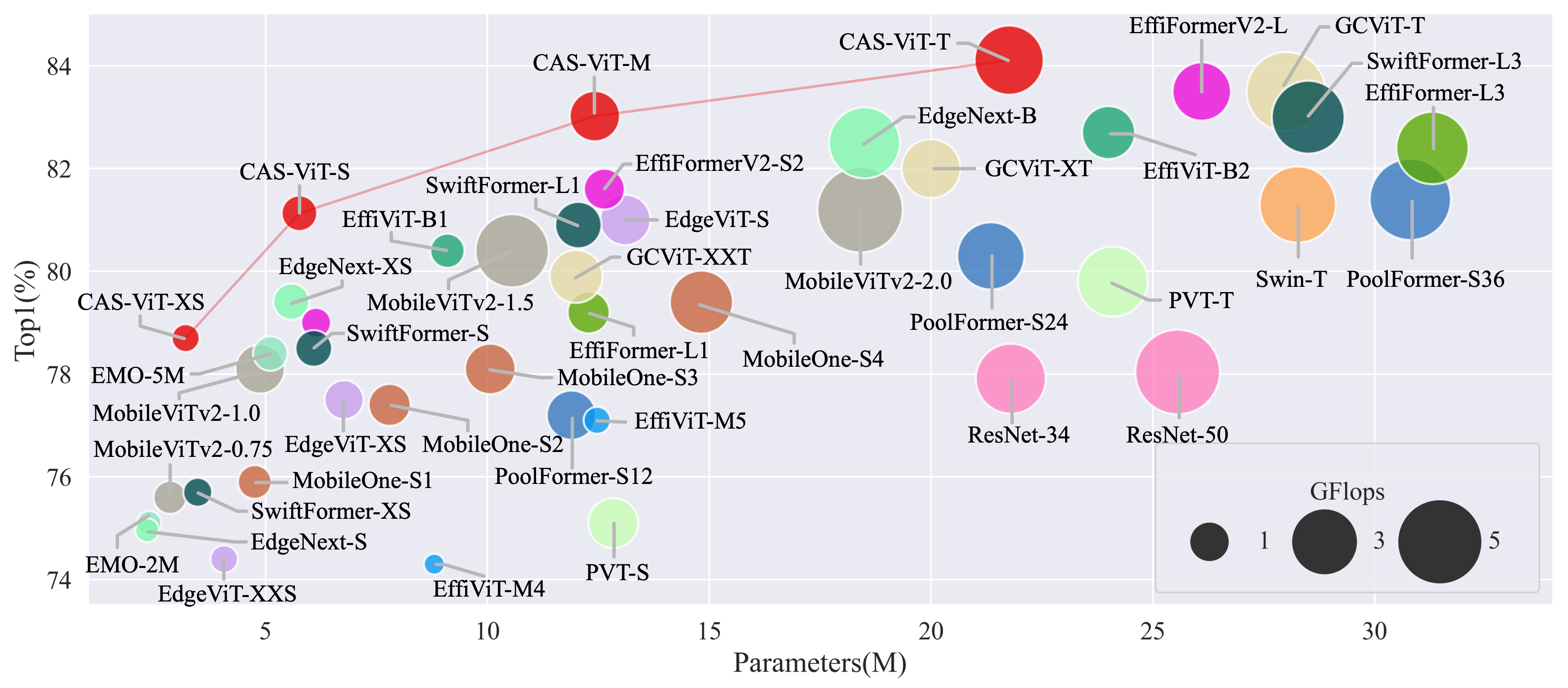}
  \caption{Parameters v.s. Top-1 accuracy on ImageNet-1K \cite{deng2009imagenet}. The circle size indicates Gflops and best viewed in color.}
  \label{fig:params_top1}
\end{figure*}

In recent years, the emergence of Vision Transformers (ViTs) marks a revolutionary shift in neural network architecture \cite{han2021transformer, huang2023vision, zhang2023attention}.
Compared to Convolutional Neural Networks (CNNs), which make a splash with lower computational complexity and higher inference efficiency \cite{zhang2022lr, zhang2023optimization}, ViT provides advantages in capturing long-range dependencies, global modeling and representation.
The basic module of ViT consists of \textit{token mixer} \cite{yu2022metaformer, yu2023metaformer}, MLP, and corresponding skip connections. In practice, \textit{token mixer} is widely implemented as Multi-head Self-Attention (MSA), which operates on the entire input sequence, compensating for the limitations of CNNs with restricted receptive fields, and providing unique advantages in terms of model scalability and adaptability \cite{wang2018non, li2023mask, wu2024rpcanet}. 

Nonetheless, deploying ViT models to mobile and edge devices remains a challenging task. 
Specifically, MSA is of quadratic complexity with respect to input image size \cite{wu2021cvt, liu2021swin} leads to computational inefficiency that fails to meet the application requirements.
Meanwhile, the complex matrix multiplication and Softmax operators in ViT suffer from operator unsupport and model conversion failure in deploying to edge devices \cite{mehta2021mobilevit}. 
As a result, on resource-constrained device deployment tasks or real-time applications, CNN is still the mainstream by virtue of its efficient computing and mature development support \cite{zhang2021agpcnet}. 
However, the restricted receptive field of CNN exposes problems such as insufficient feature extraction and underpowered performance. 
Therefore, developing \textit{token mixers} that combine both efficient deployment and powerful performance in neural networks e.g. hybrid models, has become an urgent problem on mobile devices.



Efforts have been made to improve token mixers, including MSA refinements and Heterogeneous-MSA (H-MSA). 
MSA refinements follow matrix multiplication of $\textbf{Q}$uery and $\textbf{K}$ey. It emphasizes enhancing the ability to capture long-range dependencies under the self-attention mechanism while reducing the algorithm complexity. 
Specific techniques include feature shift \cite{liu2021swin}, carrier token \cite{hatamizadeh2023fastervit}, sparse attention \cite{wang2022pvt}, linear attention \cite{han2023flatten}, etc. 
Furthermore, H-MSA is an evolutionary variant that breaks the restriction of matrix multiplication of $\textbf{Q}$ and $\textbf{K}$ to explore more flexible network designs \cite{yu2022metaformer}.
The recently proposed pool token mixer \cite{yu2022metaformer} and context vector \cite{sandler2018mobilenetv2, shaker2023swiftformer} further accelerate the inference efficiency.

Although considerable progress has been made in recent works,  
they still suffer from the following constraints: 1. High complexity associated with matrix operators such as Softmax and multiplication in token mixers limits efficient inference, while remaining problematic for mobile deployments. 
2. Networks composed of low-complexity operators fail to meet the increasing demand for powerful performance on real-time devices. The efforts still fail to achieve a satisfactory balance between efficiency and accuracy.

In this paper, we propose a family of lightweight networks named as Convolutional Additive Self-attention (CAS)-ViT to address the encountered problems. 
In particular, we argue that the capability of token mixers to obtain global contextual information hinges on multiple information interactions, such as spatial and channel domains. 
Further, by constructing an easy-to-implement additive similarity function, we eliminate computationally expensive operations and achieve self-attentive convolutional substitution.
The main contributions of our work are summarized as follows:

%


\begin{enumerate}
    
    \item We propose Convolutional Additive Token Mixer (CATM) employing underlying spatial and channel attention as novel interaction forms. This module eliminates troublesome complex operations such as matrix multiplication and Softmax.

    \item We propose Convolutional Additive Self-attention(CAS) block hybrid architecture and utilize an H-MSA structure known as CATM for each block. Further we build a family of lightweight networks called CAS-ViT, which is a powerful backbone and can be extended to various downstream tasks.
    


    \item We evaluate our method on various vision tasks, subsequently deploy it on mobile devices and report throughput data. Our XXS model achieves 78.7\% top-1 on ImageNet-1K and 14.6 throughput on iPhone. Moreover, the Tiny model achieves 84.1\% top-1 with only 21M parameters, as shown in Fig.\ref{fig:params_top1}. Our models outperform recent popular works like SwiftFormer \cite{shaker2023swiftformer}, MobileViTv2 \cite{mehta2022separable}. Extensive experiments demonstrate that our approach achieves a better balance of performance, efficient inference and easy-to-deploy.
\end{enumerate}

\section{Related Work}

\subsection{Efficient Vision Transformers}

In recent years, ViTs have represented a pivotal shift in computer vision, and its research has shown a remarkable evolution \cite{dong2022cswin, pmlr-v139-touvron21a}. 
Since the debut of ViT \cite{dosovitskiy2020image} and through its successful validation in image classification tasks on large-scale datasets such as ImageNet \cite{deng2009imagenet}, it has demonstrated the potential of self-attention mechanisms for applications in computer vision \cite{patro2024spectral, yao2024spike}. 
However, the concomitant increase in network sizes poses a huge challenge for resource-constrained scenarios such as mobile devices and real-time applications. 
To promote the potential value of ViT, workers have invested a considerable effort in efficient ViTs \cite{fan2024lightweight, shen2021efficient, liu2023efficientvit, guibas2021efficient}.

The improvement ideas for ViT architecture cover several aspects. One of which is to refine the token mixer to enhance performance or solve the quadratic complexity problem of the self-attention mechanism. For example, PVT \cite{wang2021pyramid} adopts spatial reduction strategy to achieve sparse attention to deal with high-resolution images, and Swin \cite{liu2021swin} employs a window-splitting approach to achieve local self-attention and deals with dependencies between patches by window shifting.

Another perspective is to explore hybrid models combining CNN and Transformer as compensation for the limitation of the self-attention mechanism in processing local information \cite{guo2022cmt}. EdgeViT \cite{pan2022edgevits} adopts convolutional layers and sparse attention for information integration and propagation in block respectively. NextViT \cite{li2022next} confirms the validity of hybrid models and the design strategy through comprehensive experiments. EfficientViT \cite{liu2023efficientvit} further analyzes the time-consuming proportion of various operations to achieve efficient inference.
Efficient ViTs have demonstrated exceptional versatility, excelling not only in image classification but also in a wide range of visual tasks, laying the foundation for future development.

\begin{figure*}[t]
  \centering
  \includegraphics[width=0.9\linewidth]{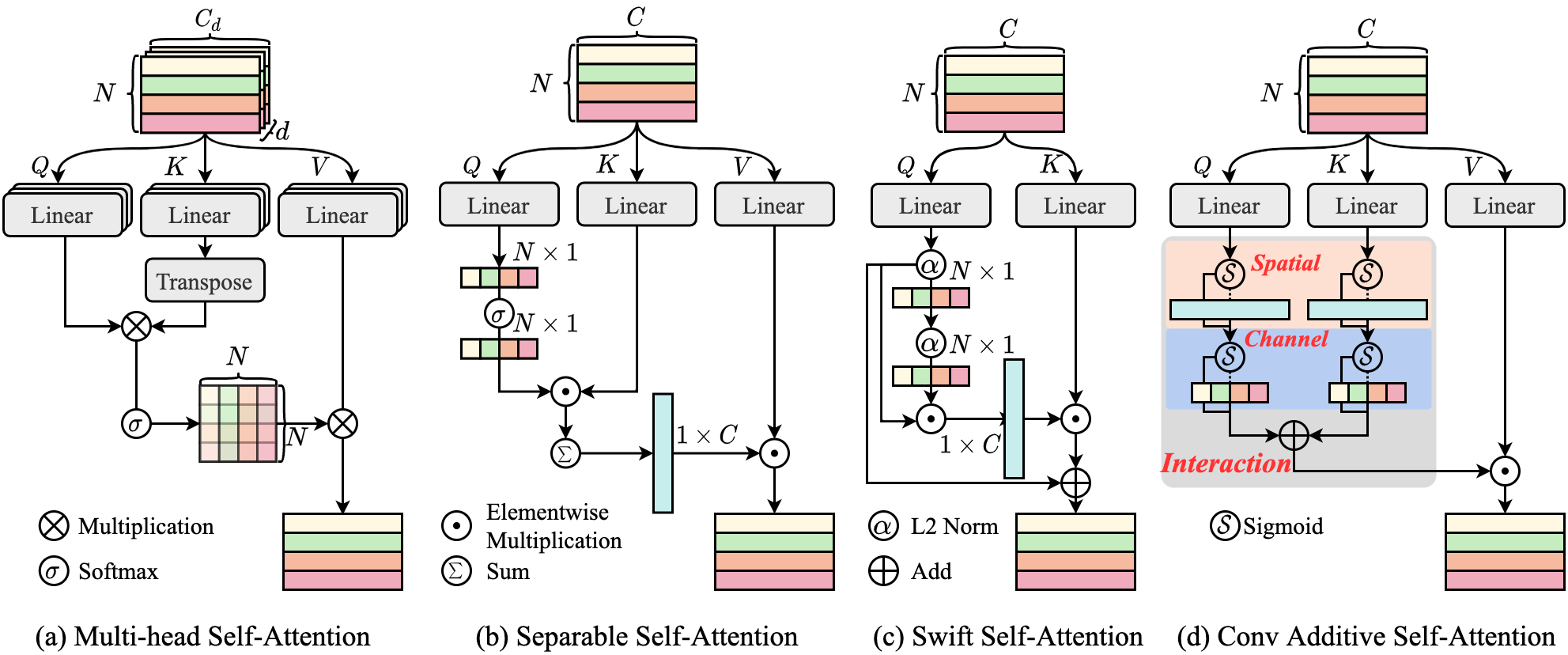}
  \caption{Comparison of diverse self-attention mechanisms. (a) is the classical multi-head self-attention in ViT \cite{dosovitskiy2020image}. (b) is the separable self-attention in MobileViTv2 \cite{mehta2022separable}, which reduces the feature metric of a matrix to a vector. (c) is the swift self-attention in SwiftFormer \cite{shaker2023swiftformer}, which achieves efficient feature association only with $\textbf{Q}$ and $\textbf{K}$. (d) is proposed convolutional additive self-attention.}
  \label{fig:tokenmixer}
\end{figure*} 

\subsection{Efficient Token Mixer}

Efficient token mixer design is one of the critical orientations in the evolution of Transformers \cite{yu2022metaformer, yu2023metaformer}.
Efforts have been made to pursue lighter and more computationally efficient token mixers to advance the feasibility of training and practical applications \cite{zhu2023biformer, li2024cpseg, patro2024scattering}.

One part of the effort is devoted to refining MSA by following the principles of $\textbf{Q}$ and $\textbf{K}$ matrix multiplication. Twins \cite{chu2021twins} introduce self-attention between patches to achieve global and local dependency tandem. Works \cite{ali2021xcit, ding2022davit, maaz2022edgenext} focus on information integration between channels and achieve self-attention between distinct domains through concatenation and transposition. 
Linear attention \cite{shen2021efficient} addresses the quadratic complexity of the self-attention mechanism by assuming that the similarity function is linearly differentiable. \cite{han2023flatten} promotes ReLU-based linear attention to be more discretized by focused function. \cite{bolya2022hydra} further simplifies the attention complexity by increasing the attention heads.

Heteromorphic-MSA (H-MSA) is an extension form of the self-attention mechanism development, 
which streamlines the multiply operations in MSA framework and intended to obtain robust features and more efficient inference performance.
Initially, MetaFormer \cite{yu2022metaformer} posited that token mixer is not the key component that affects the performance of Transformers, however PoolFormer could not be validated to be highly efficient. Subsequently, MobileViTv2 \cite{mehta2022separable} simplifies the complex matrix multiplication by endowing global information to the context vector. 
SwiftFormer \cite{shaker2023swiftformer} even eliminates 'Value' and implements weighted sums between features with simpler normalization operations, achieving a more concise and efficient H-MSA.
Inspired by previous excellent work, we rethink the token mixer design and argue that the facilitation requires information interaction on spatial and channel domains. Detailed work is described as follows.

\section{Methods}

In this section, we first review the principles of MSA and its variants. Subsequently, we will introduce the proposed CATM and focus on its distinctions and advantages over the traditional mechanism. 
Finally, we will describe the overall network architecture of CAS-ViT.

\subsection{Overview of Self-Attention and Variants}

As a crucial component of Visual Transformers, self-attention mechanism can effectively capture the relationship between different positions. Given an input $\textbf{x} \in \mathbb{R}^{N \times d}$, see Fig. \ref{fig:tokenmixer}(a), which contains $N$ tokens with $d$-dimensional embedding vectors within each head. Self-attention can be presented as follow with similarity function $Sim(\textbf{Q}, \textbf{K}) = \text{exp} \left( \textbf{Q} \textbf{K}^\top / \sqrt{d} \right)$:


\begin{equation}
    \textbf{O} = Softmax \left( \frac{\textbf{Q} \textbf{K}^\top}{\sqrt{d}} \right) \textbf{V} \enspace.
    \label{eq:self-attention-softmax}
\end{equation}


Separable self-attention \cite{sandler2018mobilenetv2}, see Fig. \ref{fig:tokenmixer}(b), reduces the matrix-based feature metric to a vector, achieving lightweight and efficient inference by decreasing the computational complexity. Subsequently, the context score is calculated through $Softmax(\cdot)$. Then, context score $\textbf{q}$ is multiplied with $\textbf{K}$ and summed along the spatial dimension to obtain the context vector that measures global information. It could be concretely described as:

\begin{equation}
    \textbf{O} = \left( \sum^N_{i=1} Softmax(\textbf{q}) \cdot \textbf{K} \right) \textbf{V} \enspace,
    \label{eq:self-attention-separable}
\end{equation}
where $\textbf{q} \in \mathbb{R}^{N \times 1}$ is obtained from \textbf{Q} key through a linear layer, $\cdot$ denotes broadcast element-wise multiplication.

Swift self-attention \cite{shaker2023swiftformer}, see Fig. \ref{fig:tokenmixer}(c), is a well-attended H-MSA architecture, which reduces the keys of self-attention to two, thereby achieving fast inference. 
It utilizes the coefficients $\alpha$ obtained by linear transformed $\textbf{Q}$ to weight each token. Subsequently, it is summed along the spatial domain and multiplied with $\textbf{K}$ to obtain the global context. It is specifically represented as:

\begin{equation}
    \textbf{O} =  \text{T} \left( \left(\sum^N_{i=1} \alpha_i \cdot \textbf{Q}_i \right) \textbf{K}\right) + \hat{\textbf{Q}} \enspace,
    \label{eq:self-attention-swift}
\end{equation}
where  $\hat{\textbf{Q}}$ denotes the normalized Query and $\text{T}(\cdot)$ is linear transformation.

\begin{figure*}[t]
  \centering
  \includegraphics[width=0.9\linewidth]{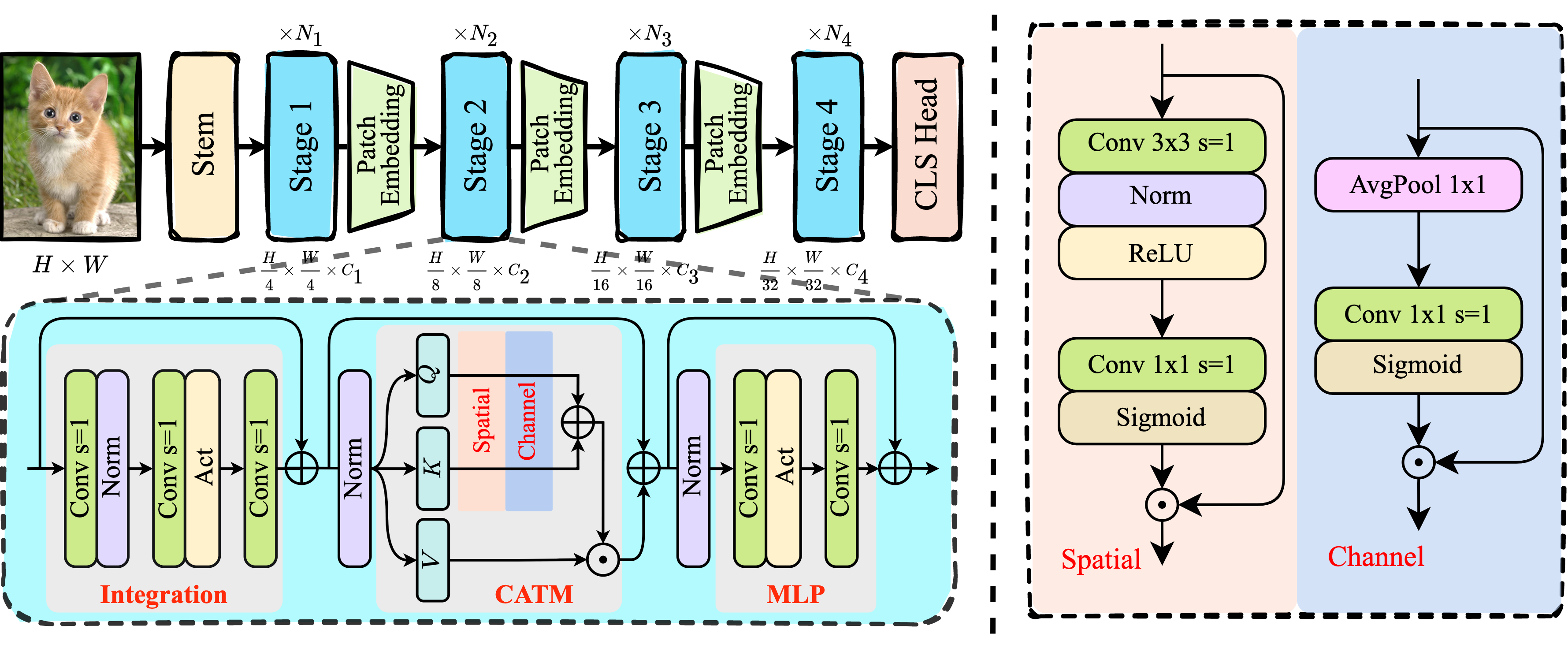}
  \caption{\textbf{Upper:} Illustration of the classification backbone. Four stages downsample the original image. \textbf{Lower:} Block architecture with $N_i$ blocks stacked in each stage.}
  \label{fig:networkarch}
\end{figure*}

\subsection{Convolutional Additive Token Mixer}
\label{subsection:catm}

\textbf{Motivation: } 
The attention graph is obtained by measuring all Query-pairs in MSA as Fig. \ref{fig:tokenmixer}(a), which leads to quadratic computational complexity $\mathcal{O}(N^2)$. 
H-MSA approaches expect to measure the correlation between tokens through a context vector as Fig. \ref{fig:tokenmixer}(b)(c).
The concept of context vector is inherited from MSA and simplified, and its generality is that it requires to perform interaction between spatial and channel domains, which can be confirmed by the operation of $N \times 1$ and $1 \times C$ vectors in Fig. \ref{fig:tokenmixer}.



Motivated by this, we argue that the information integration capability of self-attentive mechanisms resides in the multiple information interactions, e.g. the spatial association of MSA, DaViT \cite{ding2022davit} and Twins \cite{chu2021twins} in channels, and the compressive context vector of MobileViTv2 \cite{mehta2022separable} and SwiftFormer \cite{shaker2023swiftformer} in both dimensions as in Fig. \ref{fig:tokenmixer}. 
Alternatively, isn't it possible that simple and efficient operations can work better while satisfying multiple interactions?

\textbf{Token Mixer Design: } 
Inheriting from this postulate, we can consider simpler operators to meet the requirements, rather than getting involved in matrix multiplication that imposes an inference burden or Softmax that imposes a risk of not being deployable. 
Along this thought, for a tensor of size $H \times W \times C$, the multiple information interactions can be set to spatial ($N=H \times W$) and channel ($C$) domains. 
Indeed, other studies on scale decomposition have used multiple heads on channels to parallelize computation as well as dividing patches on spatial domain for speedup, however, these techniques are subdivided on the base dimensions for feature tensor.

We consider a spatial interaction as in Fig. \ref{fig:networkarch} (Right) with input $\textbf{x} \in \mathbb{R}^{H \times W \times C}$. The local token information is first integrated by a 3$\times$3 convolution layer, after which the dimension is reduced to 1 with a 1$\times$1 convolution layer, followed by a \textit{Sigmoid} activation as an attention map $\textbf{x}_s \in \mathbb{R}^{H \times W \times 1}$ in the spatial domain. Specifically represented as:

\begin{equation}
    \textbf{x}_{s} = Sigmoid (\mathcal{D}_{1 \times 1}(\mathcal{D}_{3 \times 3, ReLU, BN}(\textbf{x}))) \odot \textbf{x} \enspace.
    \label{eq:spatial}
\end{equation}

For channel domain interaction, we refer to SENet \cite{hu2018squeeze}, but instead of channel reduction we use 1$\times$1 convolution to integrate the information between channels:

\begin{equation}
    \textbf{x}_{c} = Sigmoid (\mathcal{D}_{1 \times 1}(\mathcal{P}_{1 \times 1}(\textbf{x}))) \odot \textbf{x} \enspace,
    \label{eq:channel}
\end{equation}
where $\mathcal{P}$ denotes the adaptive pooling and $\mathcal{D}$ is the group convolution layer, group number is set to channel number by default. Stacking these two operations yields the feature map after spatial and channel domain interactions, denoted as $\Phi (\textbf{x})$.

As shown in Fig. \ref{fig:tokenmixer}(d), we innovatively define the similarity function as the sum of context scores of $\textbf{Q},\textbf{K} \in \mathbb{R}^{N \times C}$:

\begin{equation}
    Sim \left( \textbf{Q},\textbf{K} \right) = \Phi (\textbf{Q}) + \Phi(\textbf{K}) 
    \label{eq:sim-additive}
\end{equation}
where Query, Key, and Value are obtained by independent linear transformations, e.g. $\textbf{Q} = W_q \textbf{x}$, $\textbf{K} = W_k \textbf{x}$, $\textbf{V} = W_v \textbf{x}$, $\Phi(\cdot)$ denotes context mapping function, which contains the essential information interactions. 
The advantage of this generalization is that it is not restricted to manual context design, and confers the possibility of implementing by convolutional operations. 
Performing multiple interactions on both $\textbf{Q}$ and $\textbf{K}$ is expected that the network learns and notices more valuable tokens. The two are integrated through additive operations, a linear approach that avoids the high complexity of matrix multiplication and is also effective in retaining useful information.
Thus the output of CATM can be formulated as:

\begin{equation}
    \textbf{O} = \Gamma \left( \Phi (\textbf{Q}) + \Phi(\textbf{K}) \right) \odot \textbf{V} \enspace,
    \label{eq:self-attention-cas}
\end{equation}
where $\Gamma(\cdot) \in \mathbb{R}^{N \times C}$ denotes the linear transformation for integrating the contextual information.


\textbf{Analysis of Receptive Field: }
ViT \cite{dosovitskiy2020image} would perform MSA after dividing the original image into patches to avoid high computational load, which would limit the receptive field to inside the patches e.g. 7$\times$7. There are many works devoted to generalize the receptive field to global \cite{liu2021swin}. 
In fact, a 7$\times$7 receptive field inside the patch can be realized by a simple three-layer 3$\times$3 convolutional stacking, although it is different from the correlation and receptive field of the self-attention mechanism, a similar effect can be achieved by multiple information interactions and convolutional stacking.

\textbf{Relation with Separable Self-Attention:} 
Compared to \cite{mehta2022separable}'s approach of extracting context scores only on Query, we perform similarity extraction on both Query and Key branches and retain the original feature dimensions on each branch. 
This enables greater preservation of visual sparse features and avoids information loss on 2D score vectors.

\textbf{Relation with Efficient Additive Attention:} First, on token mixer, we adopt a sigmoid-activated form of attention extraction instead of normalization. this facilitates network parallelization and deployment on mobile devices. In addition, the attention module in \cite{shaker2023swiftformer} is only applied to the last layer in each stage of the network, whereas the proposed CATM will be applied to each layer of the whole ViT architecture.

\textbf{Complexity Analysis:} 
In the concrete implementation, spatial and channel domain interactions are designed as a combination of depthwise convolution and Sigmoid activation, where the convolution kernel size was set to 3$\times$3, i.e:

\begin{equation}
    \begin{split}
        \Omega(\Phi (\textbf{Q})) &= \Omega(Q; S) + \Omega(Q; C) \\
        &=13HWC \enspace.
    \end{split}
    \label{eq:complexity_phi}
\end{equation}

The operation in CATM consists of four parts: the QKV separation convolution, $\Phi (\textbf{Q})$ and $\Phi (\textbf{K})$ and the final mapping layer $\Gamma$. CATM maintains linear complexity with respect to the input size:


\begin{equation}
    \begin{split}
        \Omega(\text{CATM}) &= \Omega(QKV) + 2 * \Omega(\Phi (\cdot)) + \Omega(\Gamma) \\
        &=3HWC + 2 * 13HWC + 9HWC \\
        &= 38HWC \enspace.
    \end{split}
    \label{eq:complexity}
\end{equation}




\subsection{Convolutional Additive Self-attention Block}
\label{subsection:cas}

CAS block serves as the basic structure of our method and is stacked in each stage. The design of CAS block is referenced to the widely validated hybrid model\cite{liu2023efficientvit, pan2022edgevits} and contains three parts with residual shortcuts: Integration subnet, CATM and MLP, as shown in Fig. \ref{fig:networkarch} (Left Lower). 

The integration sub-network consists of three 3$\times$3 depth-wise convolution layers activated by ReLU \cite{glorot2011deep}, this module increases the receptive field, integrates local information to facilitate the subsequent CATM, and acts as the positional encoding layer.
CATM is utilized in all blocks at each stage of the network, which is a simplified design that facilitates future development. Since the underlying network is implemented as a convolution, the CAS block can strike a good balance between computational efficiency and deployment.
MLP ratio was set to 4 by default.

\subsection{Network Architecture}

Fig. \ref{fig:networkarch}(Left Upper) illustrates the proposed network architecture. Input a natural image of size $H \times W \times 3$. It is downsampled to $\frac{H}{4} \times \frac{W}{4} \times C_1$ by two consecutive 3$\times$3 convolution layers with stride 2 in Stem. 

Afterward, it passes through four stage encoding layers, with Patch Embedding downsampling 2 times used between each stage, and obtained feature maps of size $\frac{H}{8} \times \frac{W}{8} \times C_2$, $\frac{H}{16} \times \frac{W}{16} \times C_3$ and $\frac{H}{32} \times \frac{W}{32} \times C_4$. $C_i, i \in \{1,2,3,4\}$ denotes the feature map channels. 
Each stage contains $N_i$ stacked blocks as shown in Fig. \ref{fig:networkarch}(Left Lower), and the feature map size remains constant.
We construct a family of lightweight ViT models by varying the number of channels $C_i$ and blocks $N_i$, and the specific parameter settings refer to Table \ref{table:parameterset}.

\begin{table}
\caption{Network configurations of CAS-ViT variants.}
\label{table:parameterset}
\centering
\tabcolsep=1.8mm
\renewcommand\arraystretch{1.4}
        \begin{tabular}{c|cc|cc}
        \toprule
        Model & Blocks $N$ & Channels $C$ & Params(M) & Flops(M) \\
        \midrule

        XS & [2, 2, 4, 2] & [48, 56, 112, 220] & 3.20 & 560 \\
        S  & [3, 3, 6, 3] & [48, 64, 128, 256] & 5.76 & 932 \\
        M  & [3, 3, 6, 3] & [64, 96, 192, 384] & 12.42 & 1887 \\
        T  & [3, 3, 6, 3] & [96, 128, 256, 512] & 21.76 & 3597 \\
        \bottomrule
        \end{tabular}
\end{table}

\begin{table}[!t]
    \renewcommand\arraystretch{1.3}
    \tabcolsep=0.5mm
    \begin{center}
    \caption{\textbf{Image classification on ImageNet-1k validation set \cite{deng2009imagenet}.} Throughput results are tested on NVIDIA V100 GPU, Intel Xeon Gold 6248R CPU @ 3.00GHz for ONNX and Apple Neural Engine (ANE) compiled by CoreMLTools on iPhone X IOS 15.2.}
    \label{table:cls}
    \resizebox{\linewidth}{!}{
    
        \begin{tabular}{l|cc|ccc|c}
        \toprule
        \multirow{2}{*}{Model} & Par.$\downarrow$ & Flops$\downarrow$ & \multicolumn{3}{c|}{Throughput(images/s)$\uparrow$} & Top-1$\uparrow$  \\
        \cline{4-6}
        & (M) & (M) & GPU & ONNX & ANE &(\%) \\

        \midrule
        \multicolumn{7}{c}{ConvNet-Based} \\
        \midrule
        ResNeXt101\cite{xie2017aggregated} & 83.46 & 15585 & 606 & 36.84 & 2.75 & 79.6 \\
        MobileNetv3-L-1.0\cite{howard2019searching} & 4.18 & 215 & 6303 & 226.38 & 36.70 & 75.2 \\ 
        MobileOne-S1 \cite{vasu2023mobileone}       & 4.76 & 830 & 4323 & 1366.80 & 20.85 & 77.4 \\
        MobileOne-S3 \cite{vasu2023mobileone}      & 10.07 & 1902 & 2899 & 652.73 & 16.16 & 80.0 \\ 
        MobileOne-S4 \cite{vasu2023mobileone}      & 14.80 & 2978 & 1468 & 125.20 & 9.70 & 81.4 \\

        \midrule
        \multicolumn{7}{c}{MSA-Based} \\
        \midrule
        
        EfficientViT-M4 \cite{liu2023efficientvit} & 8.80 & 301 & 6975 & 404.03 & - & 74.3 \\
        EfficientViT-M5 \cite{liu2023efficientvit} & 12.47 & 525 & 5104 & 277.10 & - & 77.1 \\
        EMO-2M \cite{zhang2023rethinking}           & 2.38 & 425 & 2672 & 80.11 & 19.07 & 75.1 \\
        EMO-5M \cite{zhang2023rethinking}          & 5.11 & 883 & 1971 & 53.61 & 10.43 & 78.4 \\
        Swin-T \cite{liu2021swin} & 28.27 & 4372 & 977 & - & - & 81.3 \\
        Swin-B \cite{liu2021swin} & 87.70 & 15169 & 616 & - & - & 83.5 \\
        EdgeViT-XXS \cite{pan2022edgevits}          & 4.07 & 546 & 2438 & 107.16 & 10.31 & 74.4 \\
        EdgeViT-XS \cite{pan2022edgevits}          & 6.78 & 1123 & 2044 & 85.17 & 6.77 & 77.5 \\
        ViT-B/16 \cite{dosovitskiy2020image} & 86.42 & 16864 & 705 & 18.27 & 0.93 & 77.9 \\
        PVT-S \cite{wang2021pyramid} & 24.10 & 3687 & 933 & 40.62 & 3.08 & 79.8 \\
        PVTv2-B1 \cite{wang2022pvt} & 14.00 & 2034 & 1385 & 21.91 & 3.41 & 78.7 \\
        PVTv2-B3 \cite{wang2022pvt} & 45.2 & 6916 & 561 & 3.48 & 1.18 & 83.2 \\
        EffiFormerV2-S0 \cite{li2023rethinking} & 3.59 & 400 & 1008 & 73.83 & 19.13 & 75.7 \\
        EffiFormerV2-S1 \cite{li2023rethinking} & 6.18 & 666 & 881 & 55.51 & 14.33 & 79.0 \\
        EffiFormerV2-S2 \cite{li2023rethinking} & 12.69 & 1268 & 485 & 31.36 & 7.10 & 81.6 \\
        
        GC ViT-XXT \cite{hatamizadeh2023global}     & 11.94 & 1939 & 951 & 41.45 & 2.76 & 79.9 \\
        GC ViT-XT \cite{hatamizadeh2023global} & 19.91 & 2683 & 1072 & 8.01 & 1.75 & 82.0 \\

        \midrule
        \multicolumn{7}{c}{Heterogeneous-MSA-Based} \\
        \midrule
        PoolFormer-S12 \cite{yu2022metaformer}     & 11.90 & 1813 & 2428 & 75.51 & 6.33 & 77.2 \\
        PoolFormer-S24 \cite{yu2022metaformer}      & 21.35 & 3394 & 1258 & 36.06 & 3.26 & 80.3 \\
        PoolFormer-S36 \cite{yu2022metaformer}      & 32.80 & 14620 & 853 & 17.29 & 2.16 & 81.4 \\
        MobileViT-XS \cite{mehta2021mobilevit}      & 2.31 & 706 & 2893 & 129.13 & 17.90 & 75.7 \\
        MobileViTv2-1.0 \cite{mehta2022separable}  & 4.88 & 1412 & 2429 & 91.40 & 11.52 & 78.1 \\
        MobileViTv2-1.5 \cite{mehta2022separable}   & 10.56 & 3151 & 1526 & 58.38 & 6.23 & 80.4 \\
        MobileViTv2-2.0 \cite{mehta2022separable} & 18.40 & 5577 & 974 & 9.33 & 3.23 & 81.2 \\
        SwiftFormer-XS \cite{shaker2023swiftformer} & 3.47 & 608 & 3184 & 132.69 & 17.11 & 75.7 \\
        SwiftFormer-S \cite{shaker2023swiftformer} & 6.09 & 991 & 2607 & 108.98 & 12.90 & 78.5 \\
        SwiftFormer-L1 \cite{shaker2023swiftformer} & 12.05 & 1604 & 2047 & 80.62 & 9.93 & 79.8 \\
        SwiftFormer-L3 \cite{shaker2023swiftformer} & 28.48 & 4021 & 1309 & 42.23 & 6.48 & 83.0 \\
        
        \midrule
        \multicolumn{7}{c}{CAS-ViT} \\
        \midrule        
        \CC{90} \textbf{CAS-ViT-XS} & \CC{90} \textbf{3.20} & \CC{90} \textbf{560} & \CC{90} \textbf{3251} & \CC{90} \textbf{115.30} & \CC{90} \textbf{14.63} & \CC{90} \textbf{78.3/78.7} \\
        \CC{90} \textbf{CAS-ViT-S} & \CC{90} \textbf{5.76} & \CC{90} \textbf{932} & \CC{90} \textbf{2151} & \CC{90} \textbf{73.06} & \CC{90} \textbf{9.92} & \CC{90} \textbf{80.9/81.1} \\
        \CC{90} \textbf{CAS-ViT-M} & \CC{90} \textbf{12.42} & \CC{90} \textbf{1887} & \CC{90} \textbf{1566} & \CC{90} \textbf{50.39} & \CC{90} \textbf{5.49} & \CC{90} \textbf{82.8/83.0} \\
        \CC{90} \textbf{CAS-ViT-T}  & \CC{90} \textbf{21.76} & \CC{90} \textbf{3597} & \CC{90} \textbf{1084} & \CC{90} \textbf{38.67} & \CC{90} \textbf{3.83} & \CC{90} \textbf{83.9/84.1} \\
        
        \bottomrule
        \end{tabular}
    }

    \end{center}
\end{table}

\begin{table}[!t]
    \renewcommand\arraystretch{1.3}
    \tabcolsep=0.5mm
    \begin{center}
    \caption{Comparisons on diverse robustness benchmarks. We report Top-1 accuracy on ImageNet-A \cite{hendrycks2019natural}, ImageNet-R\cite{hendrycks2021many}, ImageNet-Sketch \cite{wang2019learning}, and Area Under Precision-Recall curve (AUPR) on ImageNet-O \cite{hendrycks2019natural}. For these metrics, higher is better.}
    \label{table:cls_ood}
    \resizebox{\linewidth}{!}{
    
        \begin{tabular}{l|cc|cccc}
        \toprule
        Model & Par.(M)$\downarrow$ & Flops(M)$\downarrow$ & IN-A$\uparrow$ & IN-O$\uparrow$ & IN-R$\uparrow$ & IN-SK$\uparrow$ \\
        \midrule
        \midrule
        
        PVT-T \cite{wang2021pyramid} & 12.85 & 1864 & 7.8 & 18.2 & 33.7 & 21.3 \\
        PVTv2-B1 \cite{wang2022pvt} & 14.00 & 2034 & 14.7 & 19.3 & 41.8 & 28.9 \\
        EffiFormer-L1 \cite{li2022efficientformer} & 12.28 & 1310 & 10.5 & 22.0 & 42.5 & 30.0 \\
        PoolFormer-S12 \cite{yu2022metaformer}     & 11.90 & 1813 & 6.7 & 17.3 & 38.2 & 26.1 \\
        SwiftFormer-L1 \cite{shaker2023swiftformer} & 12.05 & 1604 & 17.1 & 21.8 & 45.4 & 32.6 \\

        \CC{90} \textbf{CAS-ViT-M} & \CC{90} \textbf{12.42} & \CC{90} \textbf{1887} & \CC{90} \textbf{22.1} & \CC{90} \textbf{23.4} & \CC{90} \textbf{45.9} & \CC{90} \textbf{33.6} \\

        \midrule

        Swin-T \cite{liu2021swin} & 28.27 & 4372 & 21.7 & 21.6 & 41.3 & 29.0 \\
        PVT-S \cite{wang2021pyramid} & 24.10 & 3687 & 17.7 & 20.4 & 40.4 & 27.6 \\
        PVTv2-B2 \cite{wang2022pvt} & 25.34 & 3886 & 27.9 & 22.3 & 45.1 & 32.8 \\
        EffiFormer-L3 \cite{li2022efficientformer} & 31.39 & 3940 & 23.0 & 24.3 & 46.5 & 34.6 \\
        EffiFormer-L7 \cite{li2022efficientformer} & 82.20 & 10187 & 29.0 & 26.4 & 48.2 & 35.8 \\
        PoolFormer-S24 \cite{yu2022metaformer}      & 21.35 & 3394 & 14.2 & 20.0 & 41.7 & 29.9 \\
        PoolFormer-S36 \cite{yu2022metaformer}      & 32.80 & 14620 & 18.5 & 20.5 & 42.4 & 31.1 \\
        SwiftFormer-L3 \cite{shaker2023swiftformer} & 28.48 & 4021 & 26.2 & 24.6 & 48.0 & 35.3 \\
        
        \CC{90} \textbf{CAS-ViT-T}  & \CC{90} \textbf{21.76} & \CC{90} \textbf{3597} & \CC{90} \textbf{29.6} & \CC{90} \textbf{25.4} & \CC{90} \textbf{48.5} & \CC{90} \textbf{36.0} \\
        \bottomrule
        \end{tabular}
    }

    \end{center}
\end{table}

\section{Experiments}

\subsection{Image Classification}

\subsubsection{Implementation Details}
ImageNet-1K \cite{deng2009imagenet} contains over 1.3 million images across 1,000 natural categories. The dataset covers a wide range of objects and scenes and has become one of the most extensively used datasets with its diversity. 
We train the network from scratch without any additional data, and use RegNetY-16GF \cite{radosavovic2020designing} as teacher model for distillation which was pretrained on ImageNet with 82.9\% top-1 accuracy.
The training strategy follows EdgeNeXt \cite{maaz2022edgenext}, all models are trained on input size of 224$\times$224 using AdamW \cite{loshchilov2018decoupled} optimizer for 300 epochs with batch size of 4096. Learning rate is set to 6$\times$10$^{-3}$ and the cosine \cite{loshchilov2016sgdr} decay schedule with 20 epochs warmup is used. Label smoothing 0.1 \cite{szegedy2016rethinking}, random resize crop, horizontal flip, RandAugment \cite{cubuk2020randaugment}, multi-scale sampler \cite{mehta2021mobilevit} are enabled, and the momentum of EMA \cite{polyak1992acceleration} is set to 0.9995 during training. 
To take full advantage of network effectiveness, we finetuned our model for another 30 epochs at a learning rate of $10^{-5}$ at 384$\times$384 resolution with batch size of 64. 
For specific training hyper-parameters settings, please refer to Table \ref{table:setting}.

We implement the classification model on TIMM \footnote{https://github.com/rwightman/pytorch-image-models} based on PyTorch \cite{paszke2019pytorch} and run on 16 V100 GPUs. 
In addition, we compile the Torch model into ONNX format and measure the throughput on V100 GPUs and Intel Xeon Gold CPU @ 3.00GHz with batch size 64, respectively. For the mobile side, we compiled it through CoreML library \footnote{https://apple.github.io/coremltools/docs-guides/} and throughput is measured on deployment to iPhone X neural engine.

\begin{table*}[t]
    \renewcommand\arraystretch{1.3}
    \begin{center}
    \tabcolsep=1.8mm
    \caption{\textbf{Object detection and instance segmentation performance on COCO val2017 \cite{deng2009imagenet}.} Flops are tested on image size $800 \times 1280$. Our results are shown in bold and achieve better performance with less computational overhead.}
    \label{table:det}
        \begin{tabular}{c|cc|cccccc|cccccc}
        \toprule
        \multirow{2}{*}{Backbone} & \multirow{2}{*}{Par.(M)} & \multirow{2}{*}{Flops(G)} & \multicolumn{6}{c|}{RetinaNet 1$\times$} & \multicolumn{6}{c}{Mask RCNN 1$\times$} \\
        \cline{4-15}
        &  &  & AP & AP$_{50}$ & AP$_{75}$ & AP$_S$ & AP$_M$ & AP$_L$ & AP$^b$ & AP$^b_{50}$ & AP$^b_{75}$ & AP$^m$ & AP$^m_{50}$ & AP$^m_{75}$ \\
        \midrule

        \CC{90} \textbf{CAS-ViT-XS} & \CC{90} \textbf{23} & \CC{90} \textbf{181} & \CC{90} \textbf{36.5} & \CC{90} \textbf{56.3} & \CC{90} \textbf{38.9} & \CC{90} \textbf{21.8} & \CC{90} \textbf{39.9} & \CC{90} \textbf{48.4} & \CC{90} \textbf{37.4} & \CC{90} \textbf{59.1} & \CC{90} \textbf{40.4 } & \CC{90} \textbf{34.9} & \CC{90} \textbf{56.2} & \CC{90} \textbf{37.0} \\
        
        \midrule
        PVT-T  & 33 & 240 & 36.7 & 56.9 & 38.9 & 22.6 & 38.8 & 50.0 & 36.7 & 59.2 & 39.3 & 35.1 & 56.7 & 37.3 \\
        EfficientFormer-L1  & 32 & 196 & - & - & - & - & - & - & 37.9 & 60.3 & 41.0 & 35.4 & 57.3 & 37.3 \\  
        ResNet-50  & 44 & 260 & 36.3 & 55.3 & 38.6 & 19.3 & 40.0 & 48.8 & 38.0 & 58.6 & 41.4 & 34.4 & 55.1 & 36.7 \\
        \CC{90} \textbf{CAS-ViT-S} & \CC{90} \textbf{25} & \CC{90} \textbf{189} & \CC{90} \textbf{38.6} & \CC{90} \textbf{59.2} & \CC{90} \textbf{41.2} & \CC{90} \textbf{24.0} & \CC{90} \textbf{42.1} & \CC{90} \textbf{51.2} & \CC{90} \textbf{39.8} & \CC{90} \textbf{61.5} & \CC{90} \textbf{43.2} & \CC{90} \textbf{36.7} & \CC{90} \textbf{58.8} & \CC{90} \textbf{39.2} \\
        \midrule
        
        PoolFormer-S36  & 51 & 272 & 39.5 & 60.5 & 41.8 & 22.5 & 42.9 & 52.4 & 41.0 & 63.1 & 44.8 & 37.7 & 60.1 & 40.0 \\
        SwiftFormer-L1  & 31 & 202 & - & - & - & - & - & - & 41.2 & 63.2 & 44.8 & 38.1 & 60.2 & 40.7 \\
        EfficientFormer-L3 & 51 & 250 & - & - & - & - & - & - & 41.4 & 63.9 & 44.7 & 38.1 & 61.0 & 40.4 \\
        ResNeXt101-32x4d  & 63 & 336 & 39.9 & 59.6 & 42.7 & 22.3 & 44.2 & 52.5 & 41.9 & 62.5 & 45.9 & 37.5 & 59.4 & 40.2 \\
        \CC{90} \textbf{CAS-ViT-M} & \CC{90} \textbf{32} & \CC{90} \textbf{208} & \CC{90} \textbf{40.9} & \CC{90} \textbf{61.7} & \CC{90} \textbf{43.9} & \CC{90} \textbf{25.2} & \CC{90} \textbf{44.6} & \CC{90} \textbf{53.0} & \CC{90} \textbf{42.3} & \CC{90} \textbf{64.4} & \CC{90} \textbf{46.7} & \CC{90} \textbf{38.9} & \CC{90} \textbf{61.3} & \CC{90} \textbf{41.7} \\
        \hline

        PVT-S  & 44 & 305 & 38.7 & 59.3 & 40.8 & 21.2 & 41.6 & 54.4 & 40.4 & 62.9 & 43.8 & 37.8 & 60.1 & 40.3 \\
        Swin-T  & 48 & 267 & 41.5 & 62.1 & 44.2 & 25.1 & 44.9 & 55.5 & 42.2 & 64.6 & 46.2 & 39.1 & 61.6 & 42.0 \\
        EfficientFormer-L7  & 101 & 378 & - & - & - & - & - & - & 42.6 & 65.1 & 46.1 & 39.0 & 62.2 & 41.7 \\
        SwiftFormer-L3  & 48 & 252 & - & - & - & - & - & - & 42.7 & 64.4 & 46.7 & 39.1 & 61.7 & 41.8 \\
        \CC{90} \textbf{CAS-ViT-T} & \CC{90} \textbf{41} & \CC{90} \textbf{244} & \CC{90} \textbf{41.9} & \CC{90} \textbf{62.8} & \CC{90} \textbf{45.0} & \CC{90} \textbf{25.1} & \CC{90} \textbf{46.1} & \CC{90} \textbf{55.1} & \CC{90} \textbf{43.5} & \CC{90} \textbf{65.3} & \CC{90} \textbf{47.5} & \CC{90} \textbf{39.6} & \CC{90} \textbf{62.3} & \CC{90} \textbf{42.2}  \\
        
        \bottomrule
        \end{tabular}
    \end{center}
\end{table*}

\begin{table}[h]
    \caption{CAS-ViT pretraining and finetuning hyper-parameter settings on ImageNet-1K \cite{deng2009imagenet}.}
    \label{table:setting}
    \resizebox{\linewidth}{!}{
    \begin{tabular}{l|c|c}
        \toprule
        dataset & \multicolumn{2}{c}{\textbf{ImageNet-1K}} \\
        variant & \multicolumn{2}{c}{CAS-ViT-XS/S/M/T} \\
        
        \midrule
        task & 224$^2$ pre-training & 384$^2$ finetuning \\
        \midrule
        batch size & 4096/4096/4096/2048 & 1024 \\
        base lr & 6e-3 & 5e-6 \\
        min lr & 1e-6 & 5e-6 \\
        lr scheduler & cosine & constant \\
        training epoch & 300 & 30 \\
        warmup epoch & 20 & None \\
        warmup scheduler & linear & None \\
        optimizer & AdamW & AdamW \\
        optimizer mome. & $\beta_1$,$\beta_2$=0.9,0.999 & $\beta_1$,$\beta_2$=0.9,0.999 \\
        \midrule
        color jitter & 0.4 & 0.4 \\
        auto-aug & rand-m9-mstd0.5-inc1 & rand-m9-mstd0.5-inc1 \\
        random-erase prob. & 0/0/0.25/0.25 & 0/0/0.25/0.25 \\
        random-erase mode & pixel & pixel \\
        mixup $\alpha$ & 0/0/0.8/0.8 & 0/0/0.8/0.8 \\
        cutmix $\alpha$ & 0/0/1.0/1.0 & 0/0/1.0/1.0 \\
        mixup prob. & 0/0/1.0/1.0 & 0/0/1.0/1.0 \\
        mixup switch prob. & 0.5 & 0.5 \\
        \midrule
        drop path rate & 0.0 & 0.0 \\
        label smoothing & 0.1 & 0.1 \\
        weight decay & 0.05 & 0.05 \\
        ema & True & True \\
        ema decay & 0.9995 & 0.9995 \\

        teacher model & regnety 160 & regnety 160 \\
        dist. mode & hard & hard \\
        dist. alpha & 0.5 & 0.5 \\
        dist. tau & 1.0 & 1.0 \\
        
        \bottomrule
        \end{tabular}

    }
\end{table}

\subsubsection{Results}

The experimental outcomes on the ImageNet-1K \cite{deng2009imagenet} dataset in Table~\ref{table:cls} distinctly illustrate the advancements our model introduces in the realm of image classification. Note that for throughput, we report data per frame on mobile device, and report the results on GPUs and ONNX with batch size of 64. Our results are shown in bold for all model variants. The data behind “/” is the result of training 450 epoches on ImageNet-1K \cite{deng2009imagenet}.

Compared to established benchmarks, our approach significantly elevates the precision of classification while adeptly managing the trade-offs between model complexity and computational demand.  
Specifically, our CAS-ViT-M/T models achieve top-1 accuracy of \textbf{82.8\%} and \textbf{83.9\%}, with data boosted to \textbf{83.0\%} and \textbf{84.1\%} after further training. Such performance surpasses the recent excellent work on heterogeneous MSA, SwiftFormer \cite{shaker2023swiftformer}, by \textbf{3.2\%} and \textbf{1.1\%}, with similar or even fewer parameter numbers. M model outperforms MobileViTv2-2.0 \cite{mehta2022separable} by 1.8\% with fewer parameters.
In comparison with MSA-based methods is that it outperforms EfficientFormerV2-S2 \cite{li2023rethinking} by 1.4\%, as well as our T model outperforms Swin-B \cite{liu2021swin} by \textbf{0.6\%} with \textbf{24\%} fewer parameters.

Meanwhile, the XS and S variants of our model also demonstrate a superior synergy between the parameter number and the computational efficiency measured by Flops. 
This achievement demonstrates that our model maintains a high level of performance even under limited computational conditions, which is a testament to its excellent architecture.

\subsubsection{Computational Efficiency}

Moreover, as shown in Table \ref{table:cls}, we deployed CAS-ViT to diverse platforms (GPU, ONNX, and ANE) and analyzed its computational efficiency with throughput. 
The robust model adaptation as well as high efficiency make it an ideal candidate for deployment in various scenarios, from mobile devices to high-end servers, thereby broadening the scope of its applicability in the field of image recognition.

Under similar parameter numbers, our S model has faster throughput and \textbf{2.7\%} higher performance than EMO-5M \cite{zhang2023rethinking} on all platforms, and our M model has \textbf{3.1\%} higher performance than GCViT-XXT \cite{hatamizadeh2023global}. 
Under similar throughput, other models tend to require more parameter numbers or lower accuracy, e.g. S model \textit{v.s.} EdgeViT-XS \cite{pan2022edgevits}, M model \textit{v.s.} PVTv2-B1 \cite{wang2022pvt}. 
Although the throughput falls somewhat short of methods that have a clear advantage in efficient inference, such as EfficientViT \cite{liu2023efficientvit} and MobileViTv2 \cite{mehta2022separable}, CAS-ViT has greater superiority in performance and thus is a bit more favorable in this tradeoff.

In summary, our models maintain a competitive edge in computational efficiency across all platforms. This is particularly evident when comparing the Flops and parameter counts, where our models achieve higher or comparable performance metrics with lower computational overheads. 


\subsubsection{Effective Robustness}

In order to implement a valid assessment in model robustness, we used several additional datasets:
ImageNet-A \cite{hendrycks2019natural} contains examples of misclassification by the ResNet \cite{he2016deep}, which contains 7500 adversarial-filtered images that can easily lead to model performance degradation. 
ImageNet-O \cite{hendrycks2019natural} is used to test the robustness of visual models to out-of-distribution samples, and its evaluation results are reported by Area Under Precision-Recall curve (AUPR).
ImageNet-R \cite{hendrycks2021many} contains 30000 images from categories such as art, cartoons, etc., and has deductive versions of 200 ImageNet-1K \cite{deng2009imagenet} classes.
ImageNet-Sketch \cite{wang2019learning} contains 50000 hand-drawn style images with 50 images of each of the 1000 ImageNet classes.


Our models are not finetuned or processed and are used directly for robustness evaluation and all models are evaluated on an image size of 224$^2$.
As shown in Table \ref{table:cls_ood}, we report comparative results of M and T models, with larger values indicating higher performance. 
On ImageNet-A our M and T models achieve 22.1\% and 29.6\%, outperforming other models with similar parameters.
On ImageNet-Sketch, our T model has similar accuracy to EfficientFormer-L7 \cite{li2022efficientformer} (0.2\% higher), while the number of parameters and computational effort is only 26.4\%/35.3\%.

\subsection{Object Detection and Instance Segmentation}

\subsubsection{Implementation Details}

Taking the pre-trained model on ImageNet-1K as backbone, RetinaNet \cite{lin2017focal} and Mask RCNN \cite{he2017mask} are integrated to evaluate the performance of our model on object detection and instance segmentation on MS COCO 2017 \cite{lin2014microsoft} dataset. Following \cite{yu2022metaformer}, we adopt 1$\times$ training strategy and the network is finetuned 12 epochs with learning rate of 2$\times$10$^{-4}$ and batch size of 32, and the AdamW \cite{loshchilov2018decoupled} optimizer is used. The training images are resized to 800 pixels on the short side and no more than 1333 pixels on the long side. The models are implemented on MMDeteciton \cite{mmdetection} codebase.

\subsubsection{Results}
In the comprehensive evaluation on COCO val2017 \cite{lin2014microsoft} dataset, several backbone models were assessed in terms of their Average Precision (AP) metrics across different scales (small, medium, large) and tasks (detection and segmentation), alongside their computational efficiency indicated by parameters (Par.) and Flops. 

As shown in Table \ref{table:det}, compared to ConvNet, our T model has similar number of parameters as ResNet-50 \cite{he2016deep}, and we achieve a huge advantage of \textbf{5.6} on AP box with RetinaNet as well as \textbf{5.5}/\textbf{5.2} on AP box/mask with MaskRCNN.
Compared with MSA-based models, our T model outperforms Swin-T with fewer number of parameters, and S model surpasses EfficientFormer-L1 \cite{li2022efficientformer} by 1.9/1.3 with MaskRCNN.
With a similar parameter number, our T model achieves \textbf{43.5}/\textbf{39.6} AP box/mask with MaskRCNN, outperforming the recent excellent work SwiftFormer-L3 \cite{shaker2023swiftformer} by 0.8/0.5.
Our model exhibits excellent performance, domain adaptation, and extensibility in object detection and instance segmentation tasks.


\begin{figure*}[t]
  \centering
  \includegraphics[width=0.9\textwidth]{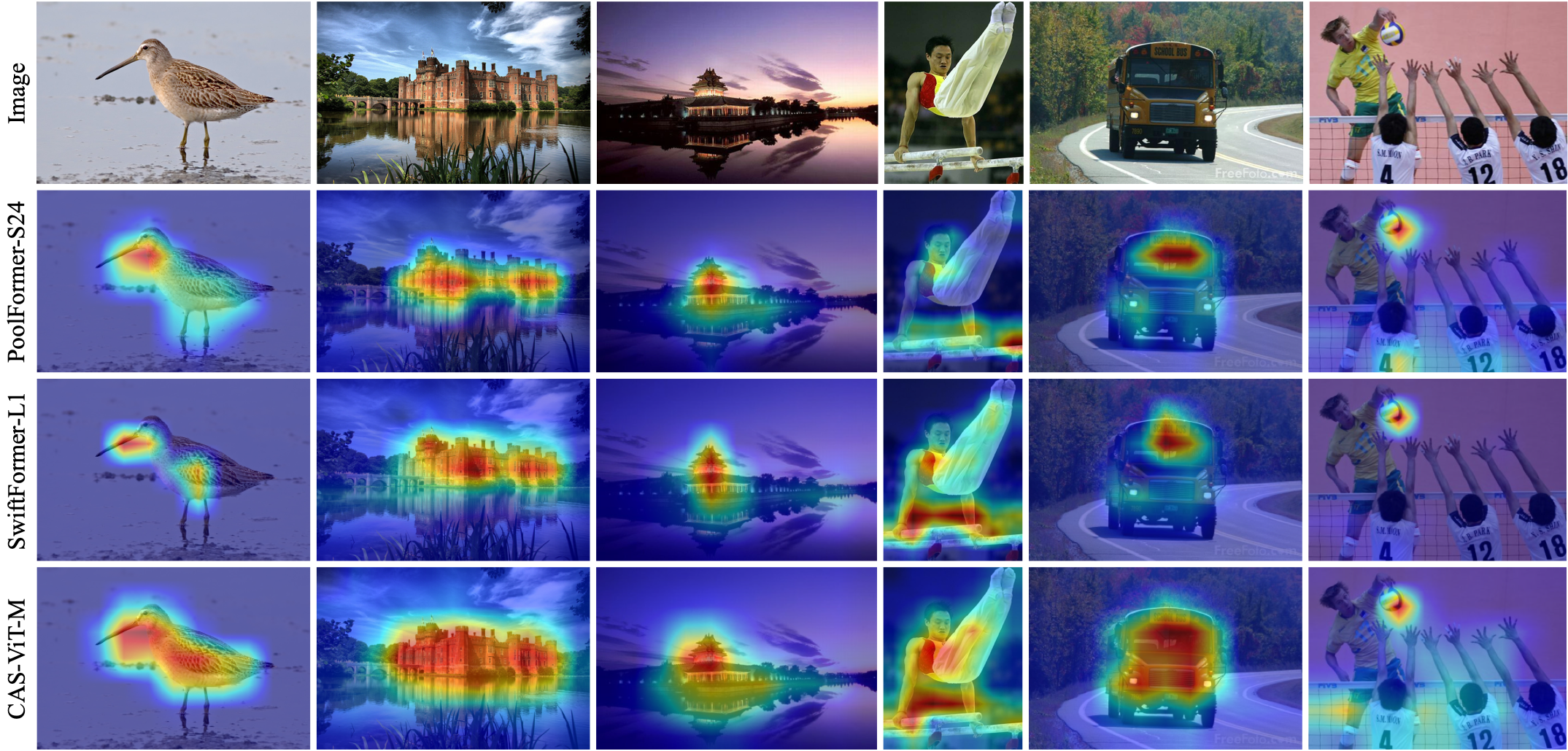}
  \caption{\textbf{Heatmap visualization for the last layer of the backbone.} From top to bottom are the original image, PoolFormer-S24 \cite{yu2022metaformer}, SwiftFormer-L1 \cite{shaker2023swiftformer} and CAS-ViT-M respectively. Our model accurately captures the global context and achieves a larger receptive field compared to other SOTAs, which is beneficial for dense prediction tasks.}
  \label{fig:heatmap}
\end{figure*}

\subsection{Semantic Segmentation}

\subsubsection{Implementation Details}

We conduct semantic segmentation experiments on ADE20K \cite{zhou2019semantic} dataset which contains 25K images covering 150 fine-grained categories.
The proposed model is evaluated with Semantic FPN \cite{kirillov2019panoptic} with the normalization layers frozen, taken as backbone, and loaded the pre-trained weights of ImageNet-1K classification.
Following the common practice \cite{yu2022metaformer}, the network is trained for 40K iterations on batch size of 32 and AdamW optimizer. The learning rate is initialized as 2$\times$10$^{-4}$ and decays in a polynomial schedule with a power of 0.9. Images are resized and cropped to 512$\times$512 during training.
The model is implemented on MMSegmentation \footnote{https://github.com/open-mmlab/mmsegmentation} codebase.



\begin{table}
\renewcommand\arraystretch{1.3}
\caption{\textbf{Semantic segmentation result on ADE20K \cite{zhou2019semantic}} combined with Semantic FPN \cite{kirillov2019panoptic}. Flops are tested on image size $512 \times 512$.}
\label{table:seg}
\centering
\tabcolsep=2.8mm
        \begin{tabular}{l|cc|c}
        \toprule
        Backbone & Par.(M) & Flops(G) & mIoU(\%) \\
        \midrule
        \CC{90} \textbf{CAS-ViT-XS} & \CC{90} \textbf{6.9} & \CC{90} \textbf{24.2} & \CC{90} \textbf{37.1} \\
        \midrule

        EdgeViT-XXS         & 7.9 & 24.4 & 37.9 \\
        EfficientFormer-L1  & 15.6 & 28.2 & 38.9 \\
        PVT-S               & 28.2 & 44.5 & 39.8 \\
        PoolFormer-S24      & 25.1 & 39.3 & 40.3 \\
        
        \CC{90} \textbf{CAS-ViT-S} & \CC{90} \textbf{9.4} & \CC{90} \textbf{26.1} & \CC{90} \textbf{41.3} \\
        \midrule

        SwiftFormer-L1      & 15.5 & 29.7 & 41.4 \\
        PoolFormer-S36      & 34.6 & 47.5 & 42.0 \\
        EfficientFormerV2-S2  & 16.3 & 27.8 & 42.4 \\ 
        PVTv2-B1            & 17.8 & 34.2 & 42.5 \\          
        
        \CC{90} \textbf{CAS-ViT-M} & \CC{90} \textbf{15.8} & \CC{90} \textbf{31.3} & \CC{90} \textbf{43.6} \\

        \midrule
        PVT-L               & 65.1 & 79.6 & 42.1 \\
        PoolFormer-M48      & 77.1 & 82.1 & 42.7 \\
        EfficientFormer-L3  & 34.7 & 42.0 & 43.5 \\
        SwiftFormer-L3      & 31.7 & 42.5 & 43.9 \\    
        XCiT-S24 P16        & 51.8 & - & 44.6 \\
        \CC{90} \textbf{CAS-ViT-T} & \CC{90} \textbf{25.0} & \CC{90} \textbf{40.39} & \CC{90} \textbf{45.0} \\
        
        \bottomrule
        \end{tabular}
\end{table}

\subsubsection{Results}
The quantitative analysis of segmentation results on ADE20K dataset, as illustrated in Table~\ref{table:seg}, reveals the superior balance between computational efficiency and segmentation accuracy achieved by our models.
Despite their relatively lower parameters and Flops, our models demonstrate competitive or superior mean Intersection over Union (mIoU) percentages, indicative of their effectiveness in semantic segmentation tasks.
Specifically, our M and T models achieve at \textbf{45.0\%} and \textbf{43.6\%} mIoU, outperforming recent work such as PVT \cite{wang2022pvt}, EfficientFormer \cite{li2022efficientformer}, SwiftFormer \cite{shaker2023swiftformer}, etc., with fewer number of parameters.
The superior performance underscores its architectural efficiency and the efficacy of its design in capturing detailed semantic information.
This balance of high efficiency and accuracy across our model variants highlights their suitability for diverse segmentation applications, especially in resource-constrained scenarios, establishing them as compelling options for efficient and effective semantic segmentation on ADE20K \cite{zhou2019semantic} dataset.

\subsection{Visualization}

\subsubsection{Heatmap}

Fig. \ref{fig:heatmap} presents a comparison of the heatmap visualizations carried out at the last layer of the networks, where the three methods share a similar parameter number. 
It can be observed that our approach is able to localize the key information in the graph very accurately, with attention to objects across animals, buildings, cars, and so on. 
Meanwhile, in comparison with two H-MSA-based methods, PoolFormer \cite{yu2022metaformer} and SwiftFormer \cite{shaker2023swiftformer}, our method is able to extract useful information over a wider range.
This also confirms our analysis on receptive fields, where the information association of feature maps is achieved in the form of convolution and does not lose the global perceptual ability to a great extent. 
In addition, a stronger perceptual capability also benefits the overall performance, especially for intensive prediction tasks such as semantic segmentation.

\begin{figure*}[!t]
  \centering
  \includegraphics[width=\linewidth]{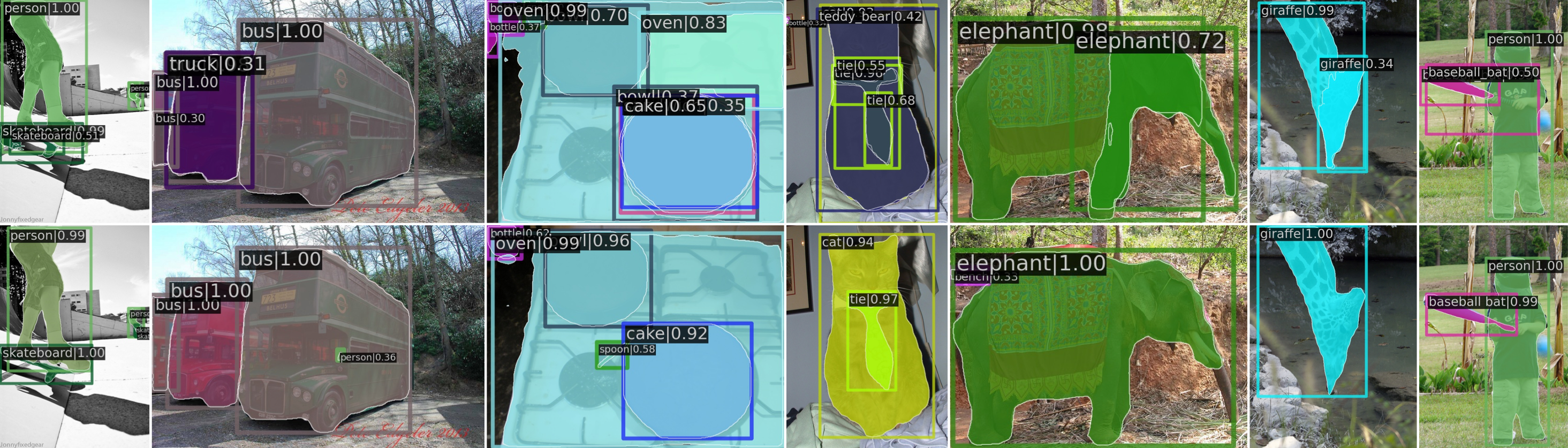}
  \caption{\textbf{Visualization of object detection and instance segmentation results on COCO 2017.} \textbf{Upper:} Prediction results of PoolFormer-S12 \cite{yu2022metaformer}, \textbf{Lower:} Prediction results of CAS-ViT-M. Our method is able to detect and segment instances accurately.}
  \label{fig:coco-vis}
\end{figure*}


\subsubsection{Detection and Segmentation Results}

Fig. \ref{fig:coco-vis} visualize the results of object detection/instance segmentation and semantic segmentation. 
Our method has the following advantages compared to PoolFormer:
1. There are fewer false detections, such as the skateboard and the cat in Fig. \ref{fig:coco-vis}; 2. There is a significant improvement in detecting objects with edges and shaded objects, e.g., the bus; 
3. For large objects, such as elephants in Fig. \ref{fig:coco-vis}, it is able to retrieve and obtain the complete bounding box in the global range.


\subsection{Ablation Study}

\begin{figure}[!t]
  \centering
  \includegraphics[width=\linewidth]{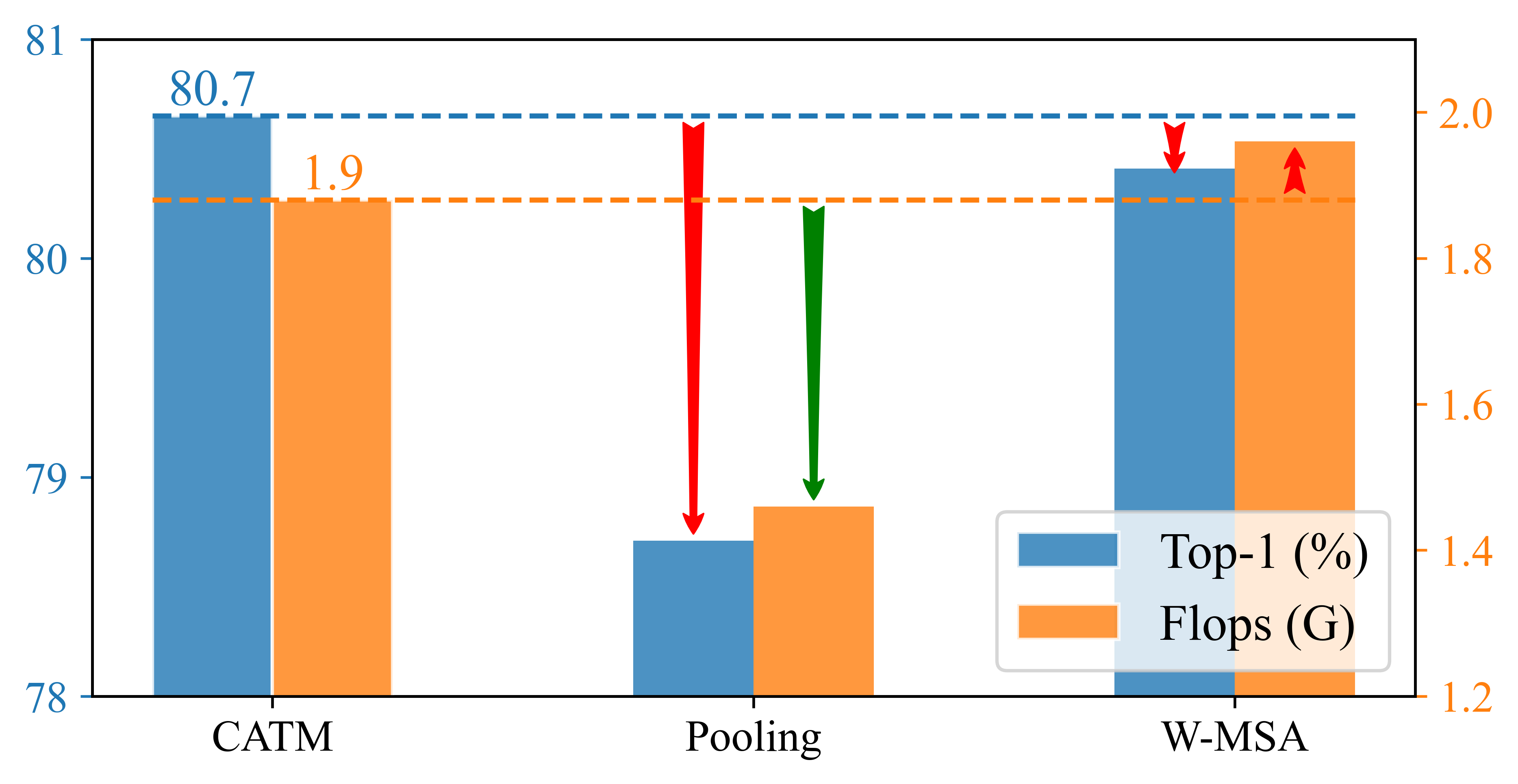}
  \caption{Ablation study on token mixer effectiveness. Experiments are conducted on ImageNet-1K \cite{deng2009imagenet}. Red arrows indicate worse indicators and green indicates better. Best viewed in color.}
  \label{fig:ablation-bar}
\end{figure}

\begin{table}[t]
    \renewcommand\arraystretch{1.4}
    \begin{center}
    \tabcolsep=1mm
    \caption{\textbf{Ablation study.} We replace CATM with other modules and replace similarity function $\Phi(\cdot)$ with other interactions.}
    \label{table:ablation}
        \begin{tabular}{clccc}
        \toprule
        $\#$ & Similarity Function & Param.(M) & Flops(G) & Top-1(\%) \\
        \midrule
        \CC{90} \textbf{1} & \CC{90} \textbf{Base} & \CC{90} \textbf{12.4} & \CC{90} \textbf{1.88} & \CC{90} \textbf{80.43} \\
        
        \midrule
        \CC{50} & \multicolumn{4}{l}{\CC{50} $Sim(\textbf{Q}, \textbf{K}) = \Phi(\textbf{Q}) + \Phi(\textbf{K})$} \\
        2 & $\Phi(\cdot)$ $\rightarrow$ w/o spatial & 12.4 & 1.86 & 79.98 \\
        3 & $\Phi(\cdot)$ $\rightarrow$ w/o channel & 11.0 & 1.88 & 80.22 \\
        \midrule
        
        \CC{50} & \multicolumn{4}{l}{\CC{50} $Sim(\textbf{Q}, \textbf{K}) = \Phi(\textbf{Q}) + \Psi(\textbf{K})$} \\
        4 & $\Phi(\cdot) = \mathcal{S}(\cdot)$, $\Psi(\cdot) = \mathcal{C}(\cdot)$ & 11.7 & 1.87 & 80.04 \\
        5 & $\Phi(\cdot) = \mathcal{S}(\mathcal{C}(\cdot))$ $\Psi(\cdot) = \mathcal{C}(\mathcal{S}(\cdot))$ & 12.4 & 1.88 & 80.40 \\

        \bottomrule
        \end{tabular}
    \end{center}
\end{table}

We demonstrate the effectiveness of the proposed method by ablation study. 
Firstly, we verify the effectiveness of CATM by replacing the token mixers. 
As shown in Fig. \ref{fig:ablation-bar}, we replace the CATM proposed in this paper with Pooling from ref \cite{yu2022metaformer} and Window-MSA. 
Pooling brings the benefit of Flops decrease and also a significant performance loss. While comparing with W-MSA, CATM is able to have higher accuracy while retaining lower Flops. It verifies the feasibility of CATM's information interaction via Sigmoid-activated attention.

Then, as shown in Table \ref{table:ablation}, we verify the arrangement of multiple information interactions in CATM to ensure that the network performance is sufficiently exploited.
Table \ref{table:ablation}(\#2-\#3) explored the effect of lacking a dimension interaction on the results, with the absence of spatial and channel decreasing by 0.45\% and 0.21\%.
 (\#4-\#5) replace the context mapping function $\Phi(\cdot)$ in the Query and Key branches with different ones, and only retaining the full interaction in \#7 achieves similar results. 
It can be concluded from this that multiple information interactions are necessary for the network to fully utilize its performance, while the arrangement order of spatial and channel domains is not particularly required.

\section{Limitation and Future Work}

CAS-ViT makes extensive usage of convolution layers for feature processing, which itself retains inductive bias. The benefits include the ability to converge faster than vanilla Transformer, and the consequent problem of being slightly less effective on large-scale datasets and large parametric models. 
In future work, we will explore on larger datasets and model scales. We are committed to developing lightweight networks that are more efficient and convenient to deploy.

\section{Conclusion}

In this paper, we proposed a convolutional additive self-attention network referred to as CAS-ViT. 
Firstly we argue that the key to enabling token mixer in ViT to work efficiently is multiply information interaction including spatial and channel domains. 
Subsequently following this paradigm, and for the network to be more readily deployed at the various edge devices, we innovatively designed an additive similarity function and implement it simply by Sigmoid-activated attention, which effectively avoids complex operations such as matrix multiplication and Softmax in vanilla ViTs. 
We constructed a family of lightweight models and verified their superior performance on tasks: image classification, object detection, instance/semantic segmentation.
Meanwhile, we have deployed the network to ONNX and Iphone, and the experiments demonstrate that CAS-ViT maintains high accuracy while facilitating efficient deployment and inference on mobile devices.

\bibliographystyle{IEEEtran}
\bibliography{ref}
 
%


 




\vfill

\end{document}